# Temporal Scale and Shift Invariant Automatic Event Recognition using the Mellin Transform

XI SHEN,[1] JULIAN GAMBOA,[1] TABASSOM HAMIDFAR,[1] SHAMIMA A. MITU, [1] AND SELIM M. SHAHRIAR[1,2]

[1]*Department of Electrical and Computer Engineering, Northwestern University, Evanston, IL 60208, USA*
[2]*Department of Physics and Astronomy, Northwestern University, Evanston, IL 60208, USA*
*\* shahriar@northwestern.edu*

**Abstract:** The Spatio-temporal holographic correlator combines the traditional 2D optical image correlation techniques with inhomogeneously broadened arrays of cold atoms to achieve 3D time-space correlation to realize automatic event recognition at an ultra-high speed. Here we propose a method to realize such event recognition for videos running at different speeds. With this method, we can highly improve recognition accuracy and filter almost all the unwanted events in the video database.

## 1. Introduction

The increasing growth of video data has generated a great demand for systems that can automatically recognize events and content. Automatic event recognition (AER) has thus emerged as a widely studied field that aims to identify specific video clips, with applications in fields from surveillance to content-based search and recommendations. In the past few years, deep learning neural networks have achieved remarkable success to accomplish these video recognition tasks [1,2]. However, the impressive accuracy of these models comes with a high computation cost that makes them difficult to integrate at large scale or in high-speed applications [2].

Optical recognition systems provide a promising alternative to the standard algorithm-oriented methods. For instance, optical image processing systems have gained significant attention in recent years due to their ability to process analog image information at ultra-high speed, substantially reducing the processing time and computational resources needed for image recognition. Numerous types of optical correlators have been developed over the decades, such as the Vander-Lugt correlator and the Joint Transform Correlator [3,4]. These optical systems are often competitive with computational techniques for certain applications thanks to their remarkable operational speed [5].

Despite this, all-optical correlators are strongly limited in that they require special materials and also cannot detect a match between otherwise identical images that have a relative rotation or scaling between them. Recently, the hybrid opto-electronic correlator (HOC) overcame the materials limitation by replacing nonlinear crystals with opto-electronic components [6–9]. In this device, input images are Fourier transformed (FT'd) optically, detected by focal plane arrays (FPAs), and then processed electronically to produce the product of the FTs. The resulting signal is then projected back out into the optical domain via a spatial light modulator (SLM) in order to be FT'd again, yielding the convolution and cross-correlation of the input images. The limitation due to rotation and scaling was then overcome by incorporating a pre-processing stage that converts rotation and scaling into linear shifts via the polar Mellin transform (PMT) [8]. Notably, the PMT can be obtained opto-electronically, as it is defined as the log-polar transform of the magnitude of the FT of an image. The FT can be obtained optically, its magnitude can be detected via an FPA, and the log-polar transform can then be performed electronically, projecting the result via an SLM. This PMT pre-processing step has been achieved at a speed of 500 frames per second (fps) [10–12], enabling shift, scale, and rotation invariant (SSRI) target recognition using images.

Unfortunately, unlike static images, videos have posed a unique challenge for optical recognition tasks due to their inherent combination of spatial features with time-domain dynamics. The implementation of an optics-based system for AER would provide a promising solution to video recognition with limited processing time and computational resources. This task goes beyond simply searching for static images within an image database, as the time-domain carries valuable information about the events captured via frame sequences.

We recently proposed a Spatio-Temporal Holographic Correlator (STHC) that combines traditional 2D optical correlation techniques with cold atoms to achieve 3D time-space correlation [13,14]. In this scheme, the atoms are inhomogeneously broadened (IHB'd) such that the spread of resonant frequencies covers the desired temporal frequency bandwidth for the system. The temporal information resulting from the interference between a plain pulse with a wide spectrum and the spectral components of the reference video are stored in the atoms in the form of coherence of different atomic states [15,16], forming a grating in the frequency domain. The spectral components of the query video diffract off these gratings to reproduce the plain pulse in case of a match, akin to the process employed for stimulated photon echoes [17,18]. Groups of these atoms can be placed in a 2D array at the Fourier plane of a lens such that each group forms a pixel that covers a small range of the spatial frequencies, thus enabling simultaneous 2D spatial and 1D temporal correlation.

While spatial SSRI can be achieved through the PMT, the playback speeds of the query and reference videos introduce a temporal scaling factor that can prevent successful matching. This is also true for the cases where otherwise identical events occur at varying speeds or are captured at different frame rates. For example, the reference event could be a video of a specific vehicle driving through a street. Regardless of the frame rate or the vehicle's speed, each occurrence should be detected. This speed or sampling variation presents a scaling problem in the time domain. The Mellin transform (MT) provides a possible solution to this problem, thus paving the way for realizing a system that can achieve speed, shift, scale, and rotation invariant ($S^3$RI) event recognition.

The rest of this paper is structured as follows. Section 2 recalls briefly the foundational model of the STHC. In Section 3, we discuss the methods and implementation details of achieving speed invariance in an STHC. Section 4 shows a comprehensive comparison of various approaches aimed at enhancing the performance of an STHC. Finally, Section 5 concludes the findings of this paper, and discusses the limitations and future work.

**2. Introduction to STHC Architecture and Implementation**

The STHC architecture has been described earlier in detail in [13]. Here, we briefly recall the basic design, as shown in Fig. 1. The applied signal sequence consists of a recording pulse, the query event frames, and the reference frames. The recording pulse is applied first, and the arrival time of the center of the pulse at the atomic medium is defined as $T_P$. Following a delay, the query event frames are sent to the SLM. The arrival time of the center of these frames at the atomic medium is defined as $T_Q$. After another delay, the reference frames are transmitted, and the arrival time of the center of these frames at the atomic medium is defined as $T_R$. The correlation signal between the query and reference event frames will be generated at the atomic medium with at the time $T_Q+T_R-T_P$. The recording pulse is chosen to be a small circle on the SLM such that its spatial FT approximates a plane wave at the atomic medium. Similarly, the duration of this pulse must be short enough to ensure that the temporal Fourier spectrum of the pulse is wider than that of the videos.

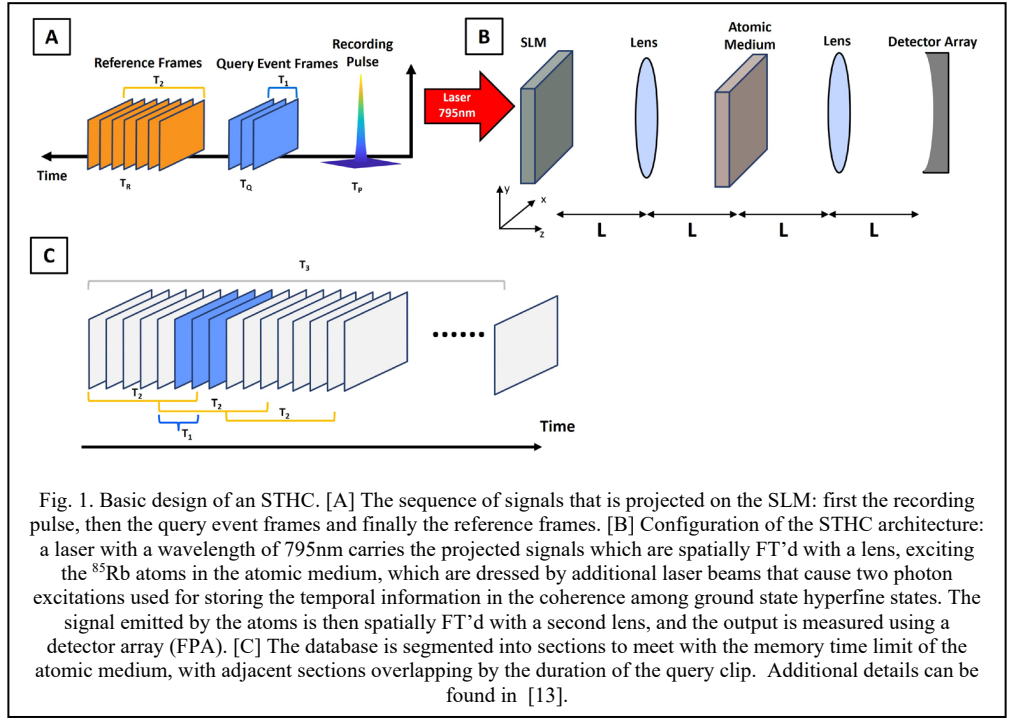

Fig. 1. Basic design of an STHC. [A] The sequence of signals that is projected on the SLM: first the recording pulse, then the query event frames and finally the reference frames. [B] Configuration of the STHC architecture: a laser with a wavelength of 795nm carries the projected signals which are spatially FT'd with a lens, exciting the $^{85}$Rb atoms in the atomic medium, which are dressed by additional laser beams that cause two photon excitations used for storing the temporal information in the coherence among ground state hyperfine states. The signal emitted by the atoms is then spatially FT'd with a second lens, and the output is measured using a detector array (FPA). [C] The database is segmented into sections to meet with the memory time limit of the atomic medium, with adjacent sections overlapping by the duration of the query clip. Additional details can be found in [13].

Since the atomic medium stores the interference between the recoding pulse and query frames in the form of a coherent superposition between the ground state and the excite state, it has a limited lifetime of the coherence time which can be around 1 second in an Rb vapor cell. However, this temporal limitation does not constrain the total length of the searchable database. As detailed in [13], the frames are loaded into the atomic system using high-speed retrieval storage media, such as holographic memory discs, allowing the practical processing time to be significantly shorter than the actual video duration. For a known input query signal and lengthy database, it is necessary to tailor the database to a optima length to correlate the query and reference video. We define $T_1$ as the loading time for the query video, and $T_2$ as the maximum duration that can be processed within the atom lifetime window. As illustrated in Fig. 1(C), the total database video length is denoted as $T_3$. The database can be segmented into multiple smaller clips, each with duration $T_2$. These segments overlap by duration $T_1$ to ensure that the AER can detect the query clip even when it spans across segment boundaries within each memory window.

Fig. 2 shows an example to illustrate qualitatively the functionality of the STHC using a series of animated frames. The query signal is shown in Fig. 2(A). The reference signal is displayed in Fig. 2(B). Here, the query frames are contained within the reference frames twice at arbitrary times labelled $t_1$ and $t_2$ with respect to the start of the video. Fig. 2(C) shows notionally that the STHC generates an output signal with high correlation peaks at times $t_1$ and $t_2$ with respect to the start of the output, corresponding to the temporal coordinate where the event was detected.

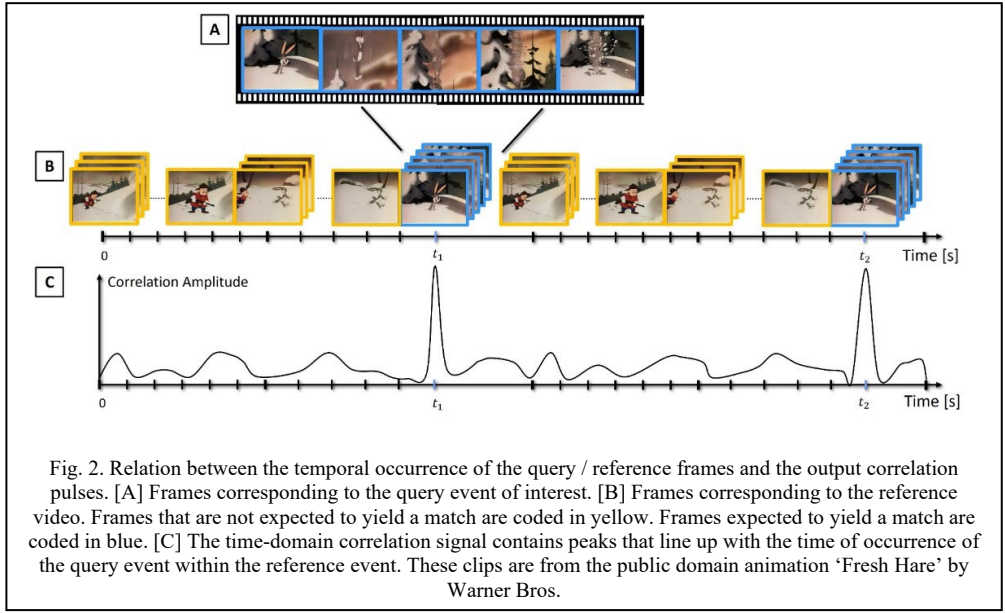

Fig. 2. Relation between the temporal occurrence of the query / reference frames and the output correlation pulses. [A] Frames corresponding to the query event of interest. [B] Frames corresponding to the reference video. Frames that are not expected to yield a match are coded in yellow. Frames expected to yield a match are coded in blue. [C] The time-domain correlation signal contains peaks that line up with the time of occurrence of the query event within the reference event. These clips are from the public domain animation 'Fresh Hare' by Warner Bros.

## 3. Speed Invariant Spatio-Temporal Correlation

The Polar Mellin Transform (PMT) is a well-established technique that can be used to perform SSRI 2D spatial correlation in opto-electronic correlators [6–11,19]. In an analogous manner, the Mellin Transform (MT) can be applied to the temporal axis of the video stream, converting temporal scaling (i.e., variations in speed) into a temporal shift and enabling the detection of events with different speeds. In order to illustrate this, we describe first the process of applying the MT for temporal correlation of a single pixel. This will be followed by a description of how this process can be extended to a 2D spatial array of pixels, thus resulting in full frame spatio-temporal correlation.

### 3.1 Speed Invariant Single-Pixel Correlation via the MT

Here two signals for a single pixel in the STHC are analyzed: the reference signal, $p_r(t)$, and the query signal, $p_q(t)$, where the latter is both shifted and scaled in the time-domain with respect to the former. We assume $p_r(t)$ vanishes for $t < 0$ and $t > T_o$, and $p_q(t)$ can be expressed as:

$$p_q(t) = p_r(\alpha \cdot [t - t_0]) \quad (1)$$

where $\alpha$ is a dimensionless scaling parameter and $t_0$ represents the relative temporal shift. The time-domain scaling is equivalent to a difference in the play-back speed and will result in an inversely proportional scaling in the Fourier spectrum.

The MT is a logarithmic coordinate transformation of the magnitude of the FT of a signal. Thus, in order to perform the MT, we must first obtain the magnitude of the FT, denoted as $\tilde{P}(\omega)$. A high-frequency cutoff is assumed implicitly, so that $\tilde{P}(\omega) = 0$ for $\omega > \omega_m$, with the value of $\omega_m$ being determined, for example, by the bandwidth of the material used for the STHC. The shift parameter $t_0$ results in a phase factor in the Fourier spectrum, which is removed when only the magnitude of the FT is used, effectively eliminating the time shift information. The process for recovering this shift parameter, if necessary, is discussed in the next section. This relation between $|\widetilde{P}_q(\omega)|$ and $|\widetilde{P}_r(\omega)|$ can be expressed as follows:

$$|\widetilde{P}_q(\omega))|=\frac{1}{\alpha}|\widetilde{P}_r(\frac{\omega}{\alpha})| \qquad (2)$$

To complete the MT, denoted $\mathcal{P}(\tau(\omega))$, the magnitude of the FT must undergo a logarithmic coordinate transformation such that the $\omega$ coordinate is mapped to $\tau(\omega) = \ln(\omega/\omega_0)$, where $\omega_0$ represents a low-frequency cut-off that is selected to represent the minimum frequency of interest in the system. Values of $\mathcal{P}(\tau)$ for $\omega < \omega_0$ are set to 0. It should be noted that $\tau(\omega/\alpha) = \tau(\omega) - \ln(\alpha)$. As such, once the MT is carried out, the scale factor $\alpha$ becomes a linear shift by the amount of $\ln(\alpha)$. The MT of the query and reference signals are thus related as follows:

$$\mathcal{P}_q(\tau) = |\widetilde{P}_q(e^\tau)| = \mathcal{P}_r(\tau - \ln(\alpha)). \qquad (3)$$

In the final step, $\tau$ functions as a new time axis. Consequently, the transformed signal retains the same information as the original data (aside from the information lost due to the high-frequency and low-frequency cut-offs), but the time scale factor is now exchanged for a linear shift factor.

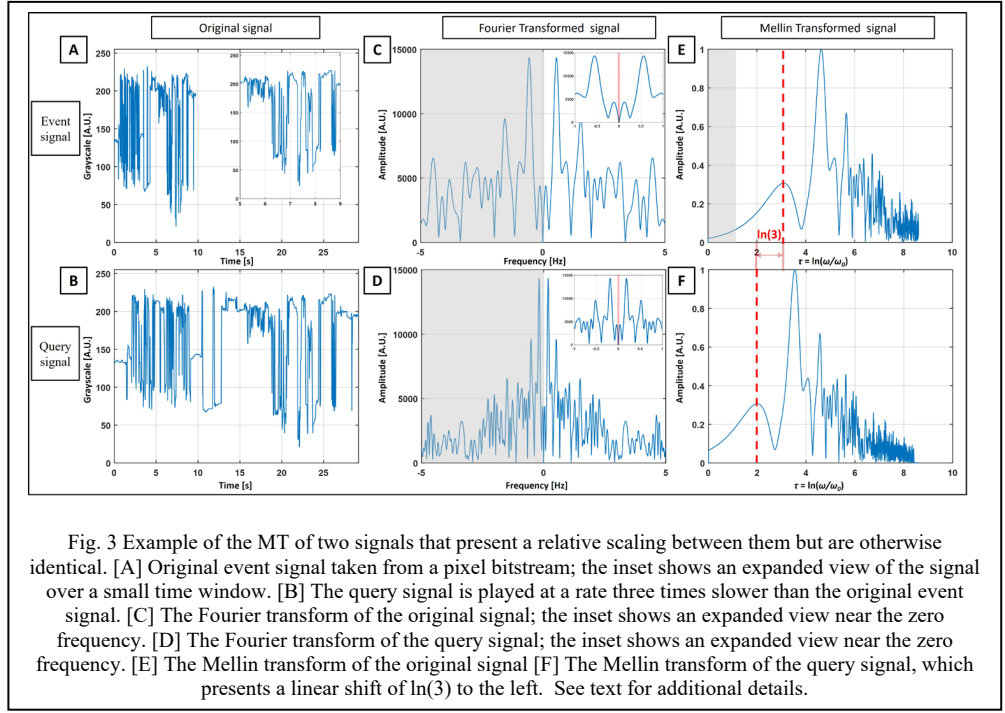

Fig. 3 Example of the MT of two signals that present a relative scaling between them but are otherwise identical. [A] Original event signal taken from a pixel bitstream; the inset shows an expanded view of the signal over a small time window. [B] The query signal is played at a rate three times slower than the original event signal. [C] The Fourier transform of the original signal; the inset shows an expanded view near the zero frequency. [D] The Fourier transform of the query signal; the inset shows an expanded view near the zero frequency. [E] The Mellin transform of the original signal [F] The Mellin transform of the query signal, which presents a linear shift of ln(3) to the left. See text for additional details.

Fig. 3 shows an example of the MT. The data shown is taken from a randomly selected pixel in the STHC example shown in Fig. 2, after the images are converted to gray scale. The original event and the query event signals are displayed in Fig. 3(A) and Fig. 3(B), respectively. The query presents the same event, albeit running 3 times slower, corresponding to a value of $\alpha = 3$. The mean amplitudes are subtracted from these temporal signals before performing the FT's, in order to suppress the irrelevant DC components. Fig. 3(C) and Fig. 3(D) show the corresponding signals after applying the FT. Since the original signals are real and non-negative, the frequency distribution is symmetrical for positive and negative frequencies. For the MT process, it is enough to consider only one side, namely that corresponding to the positive frequencies. The low frequency cut-off, chosen arbitrarily to be ~0.0045 Hz, is illustrated by the red boxes (which look like lines because of the small widths) in the insets of Fig. 3(C) and Fig. 3(D). Finally, the MT'd signals are illustrated in Fig. 3(E) and Fig. 3(F). Due to the leftward shift of the query signal, some low frequency information, shown in grey in Fig. 3(E), is lost. Given the limited amplitude and lack of distinctive features in that section of the signal, the loss of this signal is not expected to have a significant effect.

Comparison between the two MT signals in Fig. 3(E) and (F) is illustrated in Fig. 4. These signals are superimposed on each other in Fig. 4(A). This is done by shifting the MT of the event signal towards lower frequencies by an amount of ln(3), as can be seen by comparing the horizontal coordinates for the two MT plots. In Fig. 4(B), we show the difference between the MT of the event signal and the MT of the query signal, for the range over which they are expected to be identical. As can be seen, the two agree very well with each other over this range, to less than one part in a thousand.

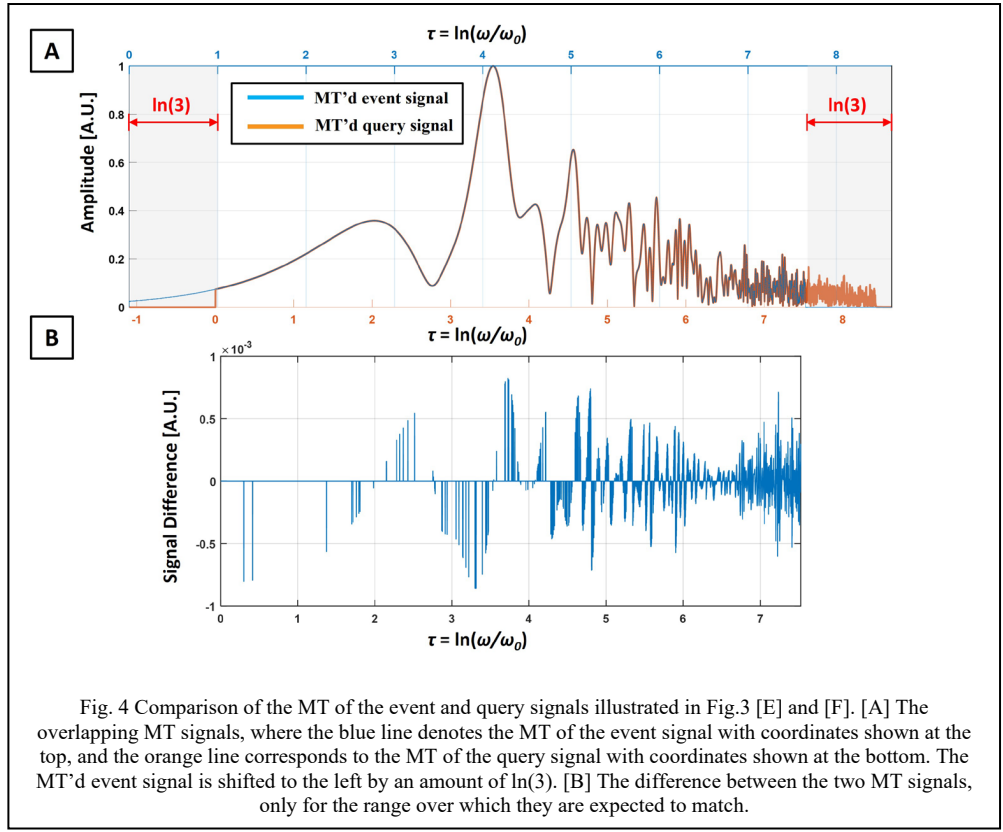

Fig. 4 Comparison of the MT of the event and query signals illustrated in Fig.3 [E] and [F]. [A] The overlapping MT signals, where the blue line denotes the MT of the event signal with coordinates shown at the top, and the orange line corresponds to the MT of the query signal with coordinates shown at the bottom. The MT'd event signal is shifted to the left by an amount of ln(3). [B] The difference between the two MT signals, only for the range over which they are expected to match.

*3.2 Recovering the Time-Shift Information in Full Frame Correlation*

In the preceding section we discussed single-pixel speed invariance through the MT, wherein any temporal scaling factor "$\alpha$" can be converted into a linear time shift factor "$\ln(\alpha)$". This technique can be extended to a full-frame array of pixels by performing each time-domain MT independently. Unfortunately, the MT process eliminates the relative time-shift that existed in the original signals, as this is encoded in the phase of the FT which is removed when taking its magnitude. Because of this, the correlation can no longer produce crucial information regarding the occurrence and timing of the events. This is problematic in AER applications, where accurately identifying events and their timing in a given video are of great importance.

We now present a two-step method (TSM) to recover the time-shift parameter while still achieving speed-invariant spatio-temporal correlation, as illustrated in Fig. 5. In Step I, we first catalog the time dependence of the signal for each pixel, for both the query video and the reference video, which are of different durations. Next, we carry out MT of the time-domain signal for each pixel, also for both videos. The MT'd videos are input into a STHC at a speed determined by the storage medium. The resulting time-domain correlation signal is mapped to the log-frequency coordinate $\tau(\omega)$. The peak amplitude indicates whether the event is detected. If there is a match detected, the peak location gives the scale factor $\ln(\alpha)$ between the query video and the event in the database.

In Step II, the query clip is adjusted by the factor obtained from Step I to align its playback speed with the events in the reference clip. These aligned videos are then input into the STHC again. The resulting correlation signal is mapped back to the original time scale, and the peak location produces the temporal position of the event in the reference video.

As discussed in the previous section, the lengthy database will be segmented into several sections, each loaded into the STHC with the query clip until the full database is searched.

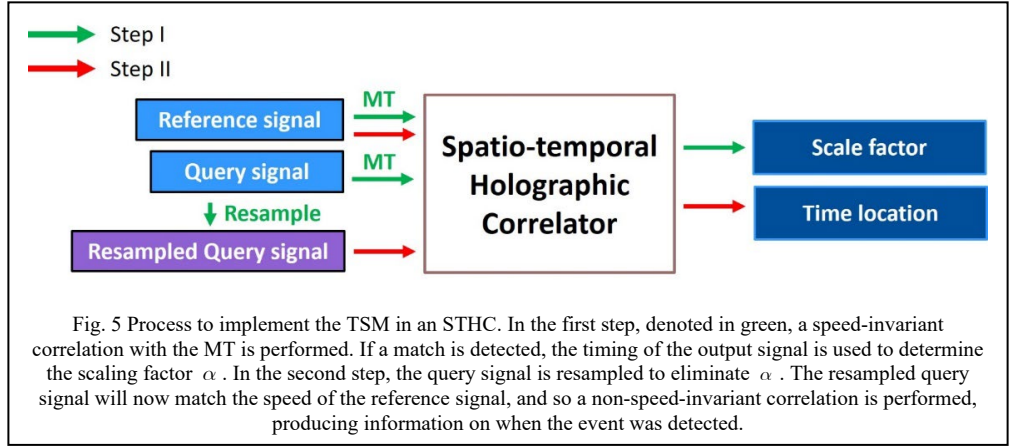

Fig. 5 Process to implement the TSM in an STHC. In the first step, denoted in green, a speed-invariant correlation with the MT is performed. If a match is detected, the timing of the output signal is used to determine the scaling factor $\alpha$. In the second step, the query signal is resampled to eliminate $\alpha$. The resampled query signal will now match the speed of the reference signal, and so a non-speed-invariant correlation is performed, producing information on when the event was detected.

A detailed process flow is shown in Fig. 6. The first step begins with a pixel-level process and subsequently combines the results of each pixel through a frame-level procedure to derive the full frame time scale factor. During the pixel-level process, each pixel bitstream undergoes the MT. Since signals played at different speeds exhibit varying frequency amplitudes, the MT signals are normalized to a maximum value of 1. Subsequently, the signals are spatio-temporally correlated. For each pixel, the output correlation signal is expressed as $c(x, y, \tau)$, where $x$ and $y$ represent the pixel's spatial coordinates, and $\tau$ is the logarithmic coordinate transformation of the temporal frequency coordinate, as explained previously. For the single-pixel case discussed in the previous section, the temporal correlation peak will occur at $\tau = \ln(\alpha)$.

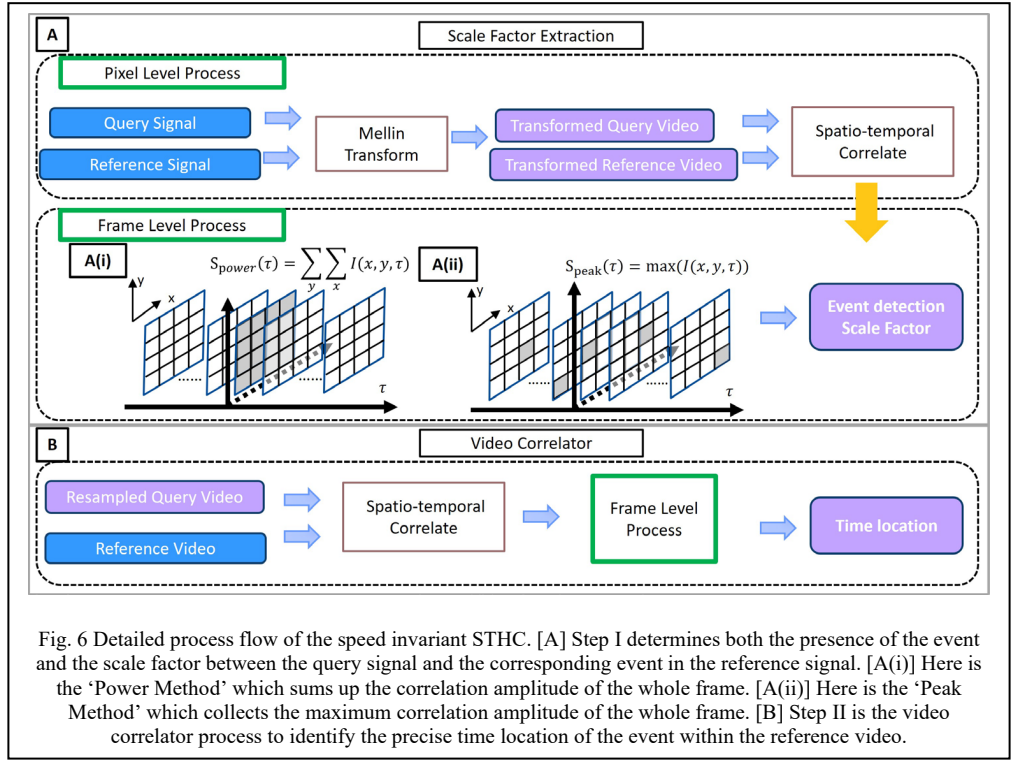

Fig. 6 Detailed process flow of the speed invariant STHC. [A] Step I determines both the presence of the event and the scale factor between the query signal and the corresponding event in the reference signal. [A(i)] Here is the 'Power Method' which sums up the correlation amplitude of the whole frame. [A(ii)] Here is the 'Peak Method' which collects the maximum correlation amplitude of the whole frame. [B] Step II is the video correlator process to identify the precise time location of the event within the reference video.

To obtain the overall temporal scaling factor for the full frame, we introduce two possible methods. The first approach is to consider the total power of the spatial correlation signal for each frame, as depicted in Fig. 6(A), and expressed as follows:

$$S_{power}(\tau) = \sum_{y_i}\sum_{x_j} c(x_j, y_i, \tau) \quad (4)$$

This cumulative result represents how the total power of the correlation peak changes as a function of time, hence it is termed the "Power Method." The second approach instead considers only the peak value of each frame in the correlation signal, as illustrated in Fig. 6(B). The expression is given by:

$$S_{peak}(\tau) = \max[c(x, y, \tau)] \quad (5)$$

The peak within each frame indicates the spatial pattern similarity between the two video signals, which can be processed before the temporal correlation. Since this method uses the spatial correlation peak amplitude to search for the temporal correlation peak location, it is termed "Peak Method".

Regardless of the selected method, the value of $\ln(\alpha)$ is measured as the time between the start of the output signal and when the maximum value of $S(\tau)$ is detected. For signals where no match is detected, the maximum value of $S(\tau)$ will remain relatively small, and so a threshold must be set such that values that surpass it will indicate a match.

After estimating the scaling factor $\alpha$ using one of the two methods described above, the speed of the query video is resampled by $1/\alpha$ such that it matches the speed of the reference. Subsequently, the adjusted videos are once again correlated albeit without the MT. The output signal determining a match has occurred will then present a peak at a time corresponding to when the event occurred.

This approach facilitates the identification and time localization of events within the video database, even when these events exhibit variations in speed. This capability is particularly advantageous for content-specific event detection, such as surveillance systems and database searching.

## 4. Temporal Scaling Factor Extraction Simulation Results

The simulations conducted in this study aimed to assess the performance of the speed invariant STHC system, with a particular focus on the temporal scaling factor extraction process. This process applies MT to achieve speed invariance, enhancing the accuracy of event detection. The precision of this process is of great importance as it directly influences the subsequent non-speed-invariant video correlation process. Here, we evaluate the efficiency of the two proposed frame-level processing methods for determining the temporal scaling factor. Finally, we analyze the performance of the two-step STHC in event detection and time determination. The frame count of the query video clips is maintained as a constant in order to more easily evaluate the accuracy of the scale factor estimation. The reference video clips include exclusively the event clips albeit at different playback speeds.

### 4.1 Performance of the Time-Scale Estimation Methods

In the previous research, we evaluated the scale factor extraction process with a single pixel bitstream [20]. Here, we evaluate the two proposed methods to achieve a full frame result. The reference video clips in the database are sourced from the large-scale and high-quality Vimeo90k dataset [21], each comprising 300 frames at a frame rate of 30 fps. 65 distinct clips are resampled at 11 different speeds to form a database containing a combined total of 715 video clips. To better test the performance of these two methods, the reference video is assumed to contain exclusively events of interest. Within the simulations, the query and reference videos are sampled equally, which corresponds to the case where they are projected on the SLM at the same frame rate. This sampling rate is not to be confused with the event speed within the videos, which has been varied as stated previously.

The scale factor, which refers to the ratio between the number of frames of the event in the reference video and that of the original event, is a crucial parameter in our analysis. To evaluate the results of each method, a standard error percentage, denoted as $\delta$, is used:

$$\delta = \frac{|\alpha_i - \alpha_t|}{\alpha_t} \times 100\% \qquad (6)$$

where $\alpha_i$ represents the estimated scale factor extracted from each video clip sample and $\alpha_t$ is the true factor. If the method yields a lower value of $\delta$, it is assumed to have higher precision due to a closer approximation of the scaling factor.

Fig. 7 shows the behavior of $\delta$ as a function of the scaling factor for each of the two methods and as a function of the query-to-reference frame ratio. In Fig. 7 the Peak Method consistently demonstrates a lower error variance across a wide range of scenarios, indicating better precision in scale factor estimation. Both methods exhibit large errors when the scale factor falls below 0.1 or exceeds 20, placing a limit on the detectable scaling factors.

It is essential to acknowledge that videos that are sped up too much (i.e., have a large scaling factor) pose a limitation due to the Nyquist–Shannon sampling theorem, which leads to information loss after performing the MT. Conversely, videos that are slowed down too much also pose a problem, albeit for a different reason. When performing the MT, a DC block is selected to avoid issues with the logarithmic coordinate transformation and reduce the impact of brightness. This cuts off low-frequency information, which disproportionately affects slow videos, causing errors in scale factor estimation. The $\omega_0$ value of the high-pass DC block can be tuned to detect a range of scaling factors based on the desired application, leading to improved outcomes. When the event is rescaled within the tuned range, minimum information is lost through the MT. In contrast, when the scale factor exceeds this range, critical frequency components may fall below the DC block or get compacted alongside higher frequencies, leading to a degradation of performance.

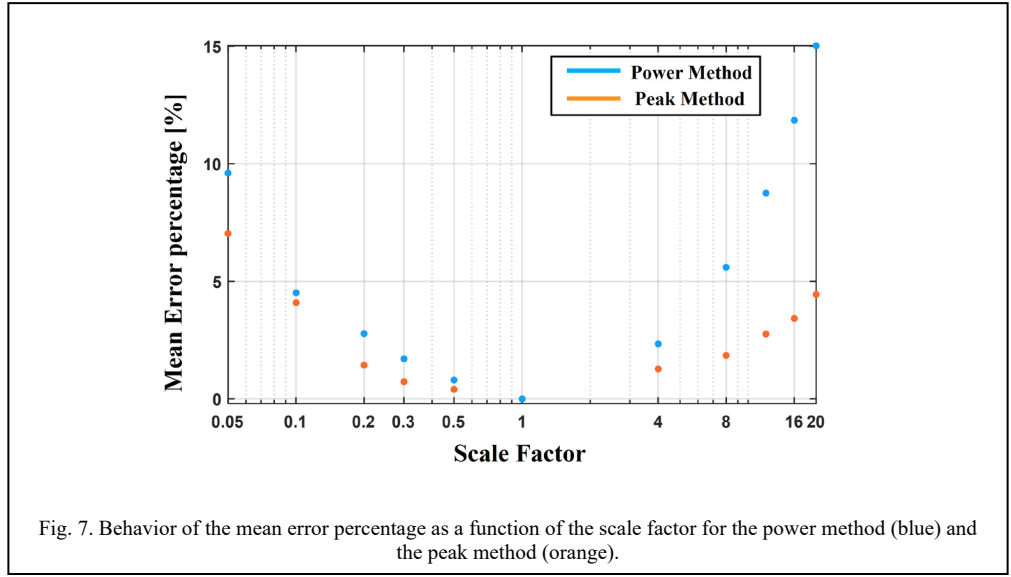

Fig. 7. Behavior of the mean error percentage as a function of the scale factor for the power method (blue) and the peak method (orange).

While the peak method yielded overall better results than the power method, it requires additional processing to identify the value of the global maxima for each correlation, and so its effectiveness is limited in scenarios requiring high-speed detection. Conversely, the power method can use a detector to quickly measure the total power of the output without additional processing.

### *4.2 Performance of the Two-Stage STHC for Determining the Occurrence Time of an Event*

The scaling factor can be obtained using an STHC with a temporal MT as described above. With this, the query video can be rescaled such that its speed matches the reference video. This allows us to pass the corrected video through a second STHC that does not incorporate the temporal MT, generating an output peak at a time that corresponds to the temporal coordinate of when the event was detected.

For the event-match simulation, we constructed a database of 50 videos derived from 10 distinct video clips, each modified with 5 different playback speeds. The 10 original clips are then individually correlated against all 50 clips in the database, resulting in 500 test pairings. We conducted two sets of simulations: one implementing the TSM as described in section 3.2, where videos underwent temporal MT for speed-invariant detection, and another without the TSM implementation.

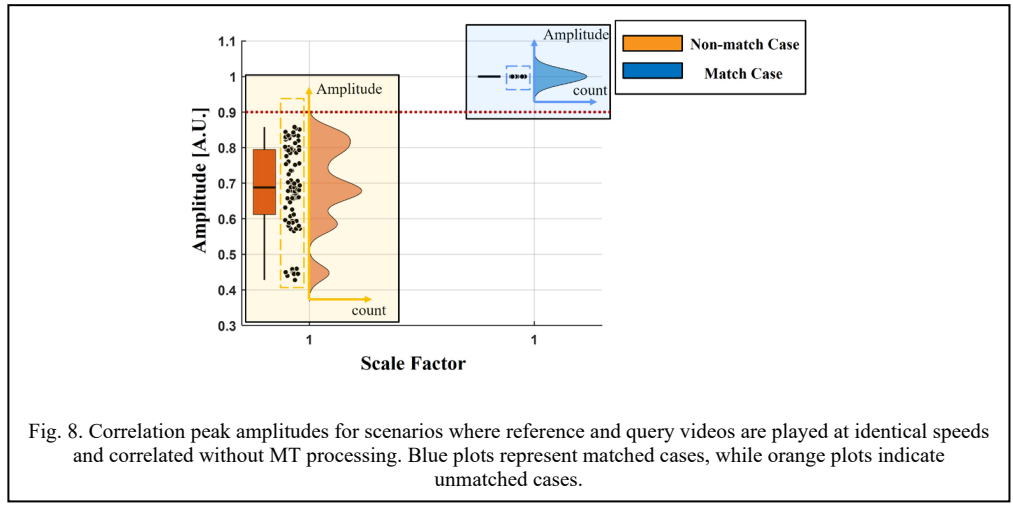

Fig. 8. Correlation peak amplitudes for scenarios where reference and query videos are played at identical speeds and correlated without MT processing. Blue plots represent matched cases, while orange plots indicate unmatched cases.

Fig. 8 provides an exemplary visualization of the correlation peak amplitude distributions for both matched and unmatched cases, displayed in blue and orange windows respectively. The correlation peaks are normalized to the peak amplitude of the auto-correlation case. The peak amplitude distribution for the non-matched case and matched case are marked in orange and blue window respectively. In each window, there are three different plots. The box plots (left) display the interquartile range, encompassing the central 50% of cases, with a central line indicating the median value of the peak amplitude. Whiskers extend from these boxes to show the full range of peak amplitudes, including maxima and minima. In the center, individual peak amplitudes from the simulation are presented as scattered dots within a dashed line window, providing an intuitive visualization of data distribution. The violin plots (right) represent the frequency distribution of peak amplitudes, with the vertical axis showing the count of cases at each amplitude level. A red threshold line is implemented as our detection criterion—peak amplitudes exceeding this threshold indicate positive matches.

Fig. 9 shows the correlation peak amplitude distributions for the simulations, with orange plots representing mismatched video pairs and blue plots denoting matched video pairs. The results without the MT and with the MT for various scaling factors are shown in Fig. 9(A) and Fig. 9(B), respectively. Fig. 9(C) shows the overall behavior without and with the MT for all scaling factors combined. In all cases, the separation between the matched and non-matched events is increased after applying the MT, enhancing detection accuracy.

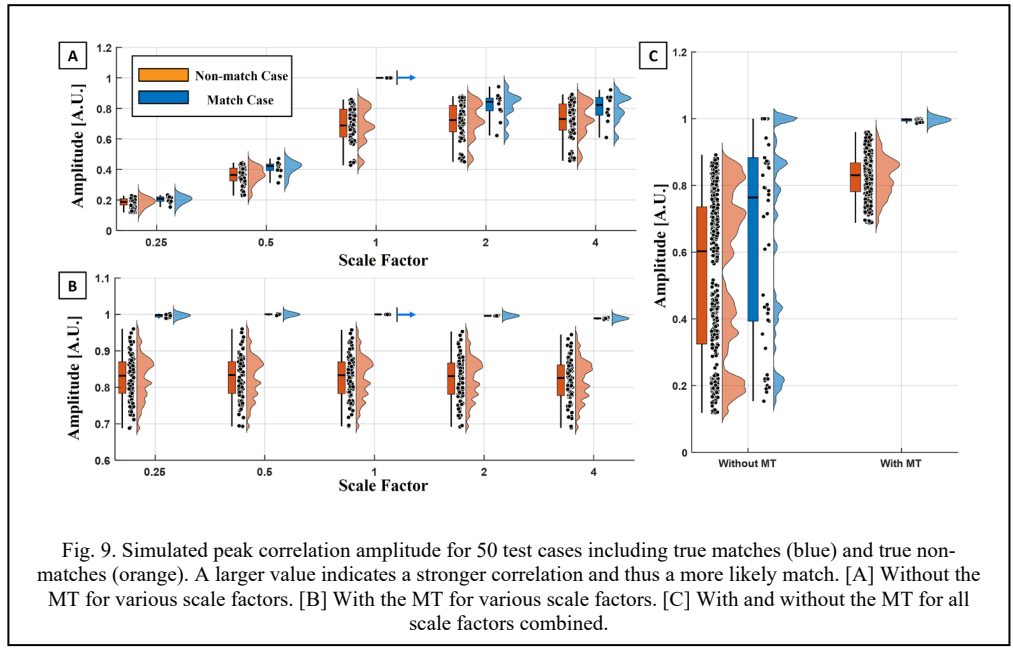

Fig. 9. Simulated peak correlation amplitude for 50 test cases including true matches (blue) and true non-matches (orange). A larger value indicates a stronger correlation and thus a more likely match. [A] Without the MT for various scale factors. [B] With the MT for various scale factors. [C] With and without the MT for all scale factors combined.

When not applying the MT, the distributions of correlation peak amplitudes for the matched and non-matched cases largely overlap, posing challenges in differentiation. Conversely, with the speed-invariant STHC approach, matched cases demonstrate noticeably higher peak amplitudes, allowing for the establishment of a solid threshold to discern matches. A threshold that minimizes false positives can be defined as the maximum measured amplitude of the non-matched test cases. Additionally, the positive detection rate is defined as the ratio of detected matches to the total number of true matches. For cases without the MT, this rate is 24%, while with the MT, it increases significantly to 100%. Alternatively, a threshold may be defined to minimize false negatives by using the minimum measured amplitude of the matched test cases. Under these conditions, the false positive rate is also 89.69% for cases without the TSM but only 0% for cases with the MT. Consequently, a substantial enhancement in detection performance is observed upon the utilization of this method. Additionally, the distribution of peak correlation amplitudes for the non-matched cases with the MT is notably concentrated. This process leads to the exclusion of all the mismatched videos, expediting the recognition process.

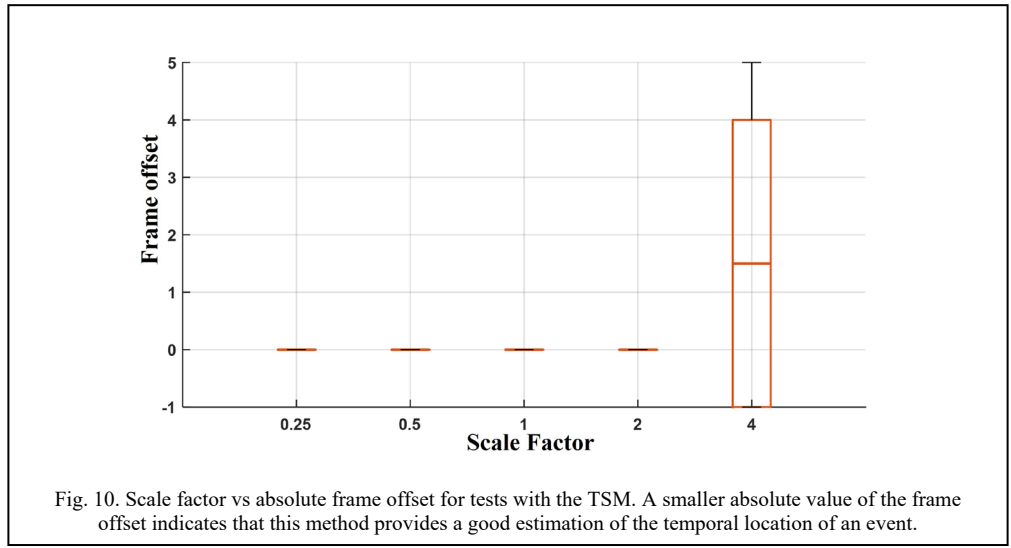

Fig. 10. Scale factor vs absolute frame offset for tests with the TSM. A smaller absolute value of the frame offset indicates that this method provides a good estimation of the temporal location of an event.

Following event-detection, Step II of TSM is applied to the matched videos to determine the precise temporal positioning. The correlation peak generated in this step can be used to find the temporal location of the video segment of interest. As a figure of merit, the frame offset can be defined for matched cases as the distance in frames between the detected temporal location and the true temporal location of the event in question. Fig. 10 shows the frame offset for tests with the TSM using various scale factors. The results demonstrate TSM's high precision in event localization: even at a scale factor of 4 with 1200 total frames, the maximum frame offset is only 5 frames, indicating exceptional temporal accuracy.

## 5. Conclusion

In this paper, we explore the utility of the MT in addressing the challenge of implementing temporal shift invariance in an AER system. To achieve full-frame video recognition, a method based on the peak value of the correlation amplitude is proposed to provide estimations for both the temporal scaling factor and location. Through an optimal sampling rate range, the TSM generates an acceptable scale factor that can be used to improve the recognition accuracy of AER.

The TSM demonstrates superior performance in event detection and ensures precise event localization. Comparative analysis reveals a 76% enhancement in detection accuracy and an 89.69% boost in detection rates compared to conventional methods, effectively filtering out all the irrelevant videos and thereby reducing the workload associated with AER tasks. Moreover, the method exhibits minimal error variance across various video types, reflecting its robustness.

### Acknowledgements

The work reported here was supported by the Air Force Office of Scientific Research under Grant Agreements No. FA9550-18-01-0359 and FA9550-23-1-0617.